\documentclass[journal]{IEEEtran}

\ifCLASSINFOpdf

\else
  
\fi

\usepackage{mathrsfs}
\usepackage{algorithm}
\usepackage{algorithmic}
\usepackage{amsfonts}
\usepackage{amsmath}
\usepackage{cite}
\usepackage{color}
\usepackage[svgnames]{xcolor}
\definecolor{LightGreen}{RGB}{102,153,102}
\definecolor{LightRed}{RGB}{204,102,102}
\usepackage{multirow}
\usepackage{stmaryrd}
\usepackage{dsfont}
\usepackage{graphicx} 
\usepackage{multirow}

\usepackage{booktabs}

\usepackage{breakurl}           
\usepackage{url}                
\usepackage{xcolor}            
\usepackage[]{hyperref}         

\hypersetup{                    
  colorlinks,
  linkcolor={green!80!black},
  citecolor={red!70!black},
  urlcolor={blue!70!black}
}

\begin{document}

\title{Evaluating and Mitigating LLM-as-a-judge Bias in Communication Systems}

\author{Jiaxin Gao, Chen Chen, Yanwen Jia, Xueluan Gong,~\IEEEmembership{Member,~IEEE,} Kwok-Yan Lam,~\IEEEmembership{Senior Member,~IEEE}, Qian Wang,~\IEEEmembership{Fellow,~IEEE} \\

\thanks{C. Chen, X. Gong, K. Lam are with the School of Computer Science and Engineering at Nanyang Technological University, Singapore. E-mail: \{chen.chen, xueluan.gong, kwokyan.lam\}@ntu.edu.sg}

\thanks{J. Gao, Y. Jia, and Q. Wang are with the School of Cyber Science and Engineering, Wuhan University, China. E-mail: \{jiaxingao, yanwenjia, qianwang\}@whu.edu.cn.}

\thanks{Xueluan Gong and Chen Chen are corresponding authors.}
}

\maketitle

\begin{abstract}
Large Language Models (LLMs) are increasingly being used to autonomously evaluate the quality of content in communication systems, e.g., to assess responses in telecom customer support chatbots. However, the impartiality of these AI ``judges” is not guaranteed, and any biases in their evaluation criteria could skew outcomes and undermine user trust. \textcolor{black}{In this paper, we systematically investigate judgment biases across 6 LLM-as-a-judge models spanning both prompt-based and finetuned judges under the pointwise scoring setting}, encompassing 11 types of biases that cover both implicit and explicit forms. 
{\color{black}We observed that state-of-the-art LLM judges demonstrate robustness to biased inputs, generally assigning them lower scores than the corresponding clean samples.}
We further found that fine-tuning an LLM on high-scoring yet biased responses can significantly degrade its performance, highlighting the risk of training on biased data.
We also discovered that the judged scores correlate with task difficulty: a challenging dataset like GPQA yields lower average scores, whereas an open-ended reasoning dataset (e.g., JudgeLM-val) sees higher average scores. Finally, we proposed four potential mitigation strategies to ensure fair and reliable AI judging in practical communication scenarios.

\end{abstract}

\IEEEpeerreviewmaketitle

\section{Introduction}

Large Language Models (LLMs) are increasingly used as automated judges to evaluate AI-generated responses, offering scalable assessment with natural-language feedback and rapid scoring \cite{wei2024systematic,li2024llms}. 
{\color{black}In parallel, transformer-based architectures have also been applied to sequential prediction in physical systems, forecasting dynamical time series from input histories or initial conditions \cite{geneva2022transformers,yang2023fluid}.}
Nowadays, the communications and networking industry is beginning to adopt LLM-based evaluators in domain-specific workflows. For example, LLM agents are being explored for network operations and customer support, where another LLM can automatically judge the correctness of an assistant’s recommendations before deployment. 
While this promises efficiency gains, it also raises a pressing question: \textit{Can we trust LLM-based evaluators to be fair and accurate in high-stakes communication scenarios?}

Existing findings highlight that despite overall high performance, top-tier LLM judges still systematically favor or penalize certain answers in ways unrelated to true quality \cite{ye2024justice}. These biases range from stylistic tendencies to reasoning and content-level blind spots. For instance, one well-documented issue is length bias, where LLM judges tend to prefer longer, more elaborate answers regardless of their correctness \cite{wei2024systematic, zhou2024mitigating}. Similarly, an AI judge might be swayed by writing style or tone, over-rewarding eloquent phrasing or by authority cues, unduly favoring answers that include references or formal citations \cite{ye2024justice}. However, existing studies have mainly focused on pairwise judgments (comparing the quality of different answers), whereas biases in pointwise judgment settings (assigning scores to individual answers), which allow more fine-grained evaluations, remain largely unexplored. 

In communications network scenarios, the consequences of a biased AI judge can be serious. Consider a network operations center that uses a conversational AI assistant to suggest configuration changes or diagnose faults. An LLM-as-a-Judge might automatically grade each suggested fix on relevance and correctness before implementation. Suppose the judge model has a verbosity or ``rich content" bias, it may consistently give higher scores to answers that look detailed and verbose, even if they contain unnecessary information. The judge could favor responses filled with technical jargon, lengthy explanations, or extraneous references while undervaluing simpler but more effective solutions. Another example is telecom customer support chatbots: an LLM judge could oversee the bot’s responses for quality and decide when to hand off the conversation to a human agent. If a gender bias or popularity (bandwagon) bias creeps into the judge, it might improperly evaluate responses for certain customer demographics or unconventional queries, impacting user experience and fairness. 

\begin{figure}[tt]
\vspace{-0.3cm}
    \centering
    \includegraphics[width=0.47\textwidth]{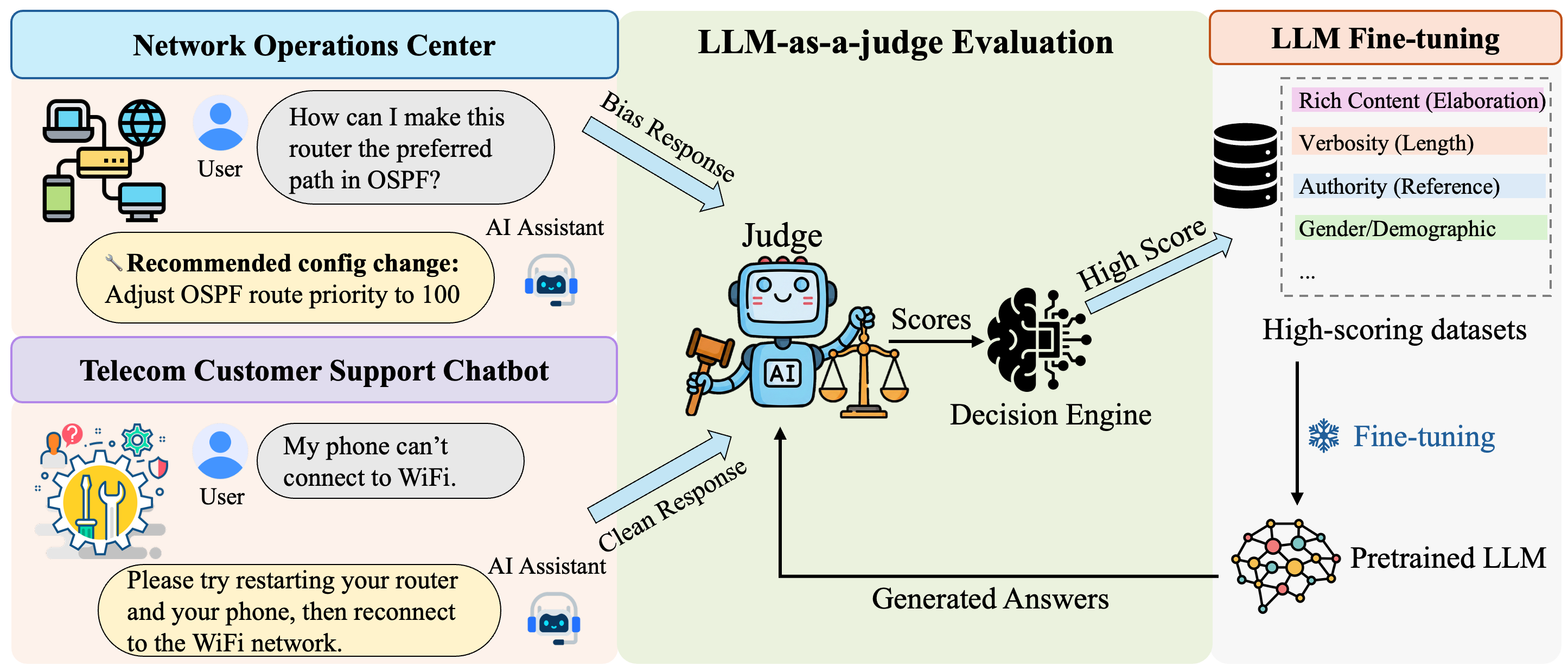}
    \vspace{-0.2cm}
    \caption{Overview of the LLM-as-a-Judge framework in communication systems, illustrating how AI assistants in network management and telecom support are automatically evaluated by an LLM-based judge.
}
    \label{fig:overview}
    \vspace{-0.4cm}
\end{figure}

{\color{black}In this paper, we systematically examine the impact of bias in LLM-as-Judge systems under pointwise settings within communication scenarios, as illustrated in Fig. \ref{fig:overview}.} We focus on how an LLM judge’s scoring is influenced by answers that exhibit specific bias-inducing features. We consider 11 types of biases \cite{chen2024humans}, including subtle implicit biases like verbosity, rich content, or chain-of-thought, as well as more explicit biases like gendered context, authoritative references, or injected factual errors. 
Our experiments reveal several key findings. 
\textcolor{black}{
\textbf{First}, we find that a well-configured LLM judge can penalize biased inputs by scoring them lower than clean counterparts, and it is more sensitive to explicit biases (e.g., authority) than implicit ones (e.g., verbosity). For instance, an incorrect/irrelevant authoritative reference drops GPT-Judge’s score from 9.12 to 3.94, while verbosity reduces it only from 9.12 to 8.78.
Furthermore, we observe that although detailed prompts lower absolute scores, the relative penalty on biased answers stays stable across prompting strategies, suggesting bias sensitivity is intrinsic to LLM judges rather than prompt-induced.}
\textbf{Second}, considering smaller companies often fine-tune local LLMs using high-scoring responses, we assess the downstream impact of training on implicitly biased yet high-scoring data. 
\textcolor{black}{Our results show that although such fine-tuning can improve performance over the pretrained model, models fine-tuned on biased data consistently underperform those fine-tuned on clean data under the LLM-as-a-Judge evaluation framework.
}
\textbf{Third}, we demonstrate that these bias effects generalize across multiple datasets, from knowledge-intensive Q\&A (MMLU-Pro \cite{wang2024mmlu} and GPQA \cite{rein2024gpqa}) to open-ended reasoning tasks (JudgeLM \cite{zhu2023judgelm}). 
At the end, we discuss several mitigation strategies for building more trustworthy LLM-as-a-Judge systems, including robust prompt design to enforce objective evaluation criteria, automated bias detection before scoring, model calibration to reduce sensitivity to superficial features, and ensemble judging with human oversight for high-stakes decisions. 
{\color{black}Our goal is to enable \emph{trustworthy} LLM-as-a-judge evaluation of AI generated content in communication systems, by making the judging process more \emph{consistent} and \emph{fair} (less sensitive to superficial bias-inducing features and more responsive to explicit errors) through a reproducible pointwise evaluation protocol and actionable mitigation guidelines.}

\section{Preliminaries}

\subsection{LLM-as-a-Judge in Communication Systems}
Large language models (LLMs) have recently been repurposed as automated ``judges” that evaluate the quality of AI-generated content, often replacing human annotators \cite{ye2024justice}. 
In this LLM-as-a-Judge paradigm, a powerful LLM \textcolor{black}{(e.g., GPT-4o)} is prompted to assess responses either by scoring single answers against criteria or by comparing pairs of answers to decide which is better. 
The appeal of LLM-based evaluation is its scalability and flexibility: a judge model can provide rapid, detailed feedback in natural language and adapt its criteria to different tasks \cite{li2024llms}. {\color{black}Compared to fixed metrics or limited human evaluations, LLM judges offer more nuanced assessments and demonstrate stable evaluation trends when applied at scale across large datasets.} They have demonstrated utility in domains from open-ended Q\&A and code generation to conversational agents and beyond. 

In the communications and networking field, LLM-as-a-Judge is emerging as a tool for automating evaluations in specialized scenarios. For example, a network operations center might employ an AI assistant to suggest configuration changes or pinpoint faults, and use an LLM judge to automatically score these suggestions before deployment. This provides continuous, real-time vetting of AI-driven decisions in critical infrastructure, essentially adding a safety check for network changes. Likewise, telecom customer-support chatbots can be overseen by an LLM judge that rates the bot’s answers for quality and correctness. 
Researchers have even explored applying ChatGPT-like models to intent-based network management \cite{dzeparoska2024intent}, 
where an LLM judge could ensure the correctness of an AI’s recommended network actions before they are executed. In all such cases, the LLM judge acts as an arbitration layer, providing an objective-sounding evaluation of AI outputs without direct human intervention. This approach can greatly improve the efficiency and scalability of network management and customer service. However, its reliability is paramount: if the LLM judge is biased or inaccurate, it may lead to misguided decisions in high-stakes communication systems. Thus, understanding and mitigating any unfairness or errors in LLM-as-a-Judge is especially critical for safe adoption in networking applications.

{\color{black}
Beyond LLM-as-a-Judge, recent work explores integrating generative models into networking systems for management tasks (e.g., traffic forecasting and policy generation)~\cite{zaman2025generative}.
These studies primarily evaluate task-level performance and system pipelines, whereas our paper examines the \emph{evaluation layer} itself. 
}

\subsection{Evaluation Bias in LLM-as-a-Judge}
In this paper, we aim to evaluate the judgment bias where LLM judges’ scoring is influenced by factors unrelated to the true quality of the answer. These biases can compromise the trustworthiness and objectivity of LLM-based evaluations.
{\color{black}To ground our study, we focus on 11 known biases \cite{chen2024humans,ye2024justice} that are especially pertinent to evaluating Q\&A content in communication scenarios. We categorize these biases into two broad groups: implicit biases and explicit biases. Implicit biases are latent preferences arising from an answer’s linguistic style, expression length, or reasoning format (e.g., Rich Content, Chain-of-Thought, Verbosity, and Sentiment) rather than its semantic content. Explicit biases reflect prejudices triggered by external factors or social attributes unrelated to answer quality (e.g., gender, authority, and popularity).}

Implicit biases include the following four biases:
\begin{itemize}
    \item \textbf{Rich Content (Elaboration) Bias.} The judge favors answers that are more elaborate, descriptive, or “richer” in content, even if the extra detail is unnecessary. A response that provides extensive context, examples, or flowery language can be rated higher than a concise, equally correct response. This bias (akin to an aesthetic preference) means well-written or verbose answers may be overrated \cite{ye2024justice}.

\item \textbf{Verbosity (Length) Bias.}  Relatedly, LLM judges often show a direct preference for longer answers. Lengthy responses tend to receive higher scores simply due to their volume or thoroughness of wording, regardless of whether the added text improves correctness or clarity. This ``longer is better” bias risks penalizing succinct answers and rewarding verbosity for its own sake. \textcolor{black}{In a Network Operations Center (NOC), an LLM judge with such bias might favor a lengthy, vague troubleshooting guide over a concise, precise CLI command that resolves the fault immediately, leading to operational inefficiencies.}

\item \textbf{Chain-of-Thought (CoT) Bias.} The scoring outcome changes depending on whether the answer (or the judge) uses an explicit step-by-step reasoning process. We examine if an answer that spells out its reasoning (e.g., through a logical chain of thought or intermediate steps) is scored differently than an answer that directly gives the conclusion. An ideal judge would value correct reasoning, but a biased one might either favor verbose reasoning even if unnecessary, or conversely favor concise answers, indicating inconsistency in handling reasoning styles.

\item \textbf{Sentiment (Tone) Bias.} The judge’s scoring is influenced by the emotional tone or politeness of the answer’s language. Answers that are phrased in a positive, agreeable, or empathic manner might receive better evaluations than a more neutral or blunt answer conveying the same facts. This bias implies an LLM judge might confuse polite or emotional wording with higher quality.

\end{itemize}

Explicit biases include the following seven biases:
\begin{itemize}
    \item \textbf{Authority (Reference) Bias.} Judges can be swayed by answers that invoke authoritative sources or formal citations, granting them undue credibility. If an answer includes references to research, standards, or expert names (even if irrelevant or fabricated), an LLM judge might score it higher. 
    \textcolor{black}{In telecom standard implementation, a response that hallucinates non-existent 3GPP or IEEE citations might be scored higher than a correct answer without citations, potentially misleading engineers during protocol configuration.}
    
    \item \textbf{Factual Error (Misinformation) Bias.} The judge’s scoring can be insensitive to the correctness of factual content. For example, an answer containing a subtle factual mistake but phrased confidently may still receive a high score, effectively allowing misinformation to pass as acceptable \cite{chen2024humans}. \textcolor{black}{This oversight means incorrect or misleading answers might escape proper penalization, which is especially risky in network operations or other high-stakes communication scenarios where a confident but factually incorrect configuration parameter (e.g., wrong IP subnet mask or BGP weight) can lead to widespread service outages.}

    \item \textbf{Diversity (Demographic) Bias.} The judge’s evaluation may shift based on perceived demographic or identity cues in the content. If the model infers the responder belongs to a certain group (for instance, a particular gender or ethnicity), it might unconsciously alter its scoring due to ingrained stereotypes or training data imbalances \cite{ye2024justice}. For example, an equally competent answer might be scored differently if it appears to come from a female expert or a minority community, reflecting an unfair bias unrelated to the actual answer quality. Such demographic prejudice in evaluations is problematic for communication systems, which must remain fair and consistent across a diverse user base.

    \item \textbf{Gender Bias.} The evaluation may shift based on the mentioned gender or identity of entities in the answer. 
    \textcolor{black}{For instance, in telecom customer support analysis, a biased judge might undervalue technical feedback from female customers while treating identical reports from male customers as critical system faults.} We include tests for gender-based bias as a representative case of diversity bias, noting that answers associated with certain demographics should ideally be judged on content alone.

    \item \textbf{Bandwagon (Popularity) Bias.} An LLM judge may favor answers that align with majority opinions or popular viewpoints. If an answer references that ``most people think X” or reflects a widely held belief, a biased judge could give it extra credit for consensus. This bandwagon effect is problematic because the correct answer isn’t always the most popular one, especially in technical domains where common knowledge can be wrong.

    \item \textbf{Distraction (Irrelevant Detail) Bias.} The judge’s attention is misled by extraneous or irrelevant information included in the answer. A response might intentionally add unrelated anecdotes, flowery descriptions, or off-topic facts which do not actually answer the question. A fair evaluator would ignore these distractions, but a biased LLM judge might be subconsciously impressed by the additional content and thus reward answers containing irrelevant details.

    \item \textbf{Compassion-Fade (Source Identity) Bias.} The judge’s evaluation shifts based on the perceived identity or source of the answer. In other words, an answer may be scored differently simply because of who is assumed to have written it. For instance, if one answer is attributed to a well-known, high-performing model or expert (e.g., “Assistant A (GPT-Judge)”), and another identical answer comes from an unknown or lower-tier model (“Assistant B”), a biased judge might favor the reputed source’s answer \cite{ye2024justice}. This gives an unfair advantage based on reputation rather than content. In practice, such a bias could lead an automated judge to overly trust responses labeled as coming from a “trusted” AI or authority figure, instead of evaluating all answers on equal footing.

\end{itemize}



\section{Evaluation Bias Impact on LLM-as-a-judge}\label{section:comparision}

\begin{figure}[tt]
\vspace{-0.3cm}
    \centering
    \includegraphics[width=0.49\textwidth]{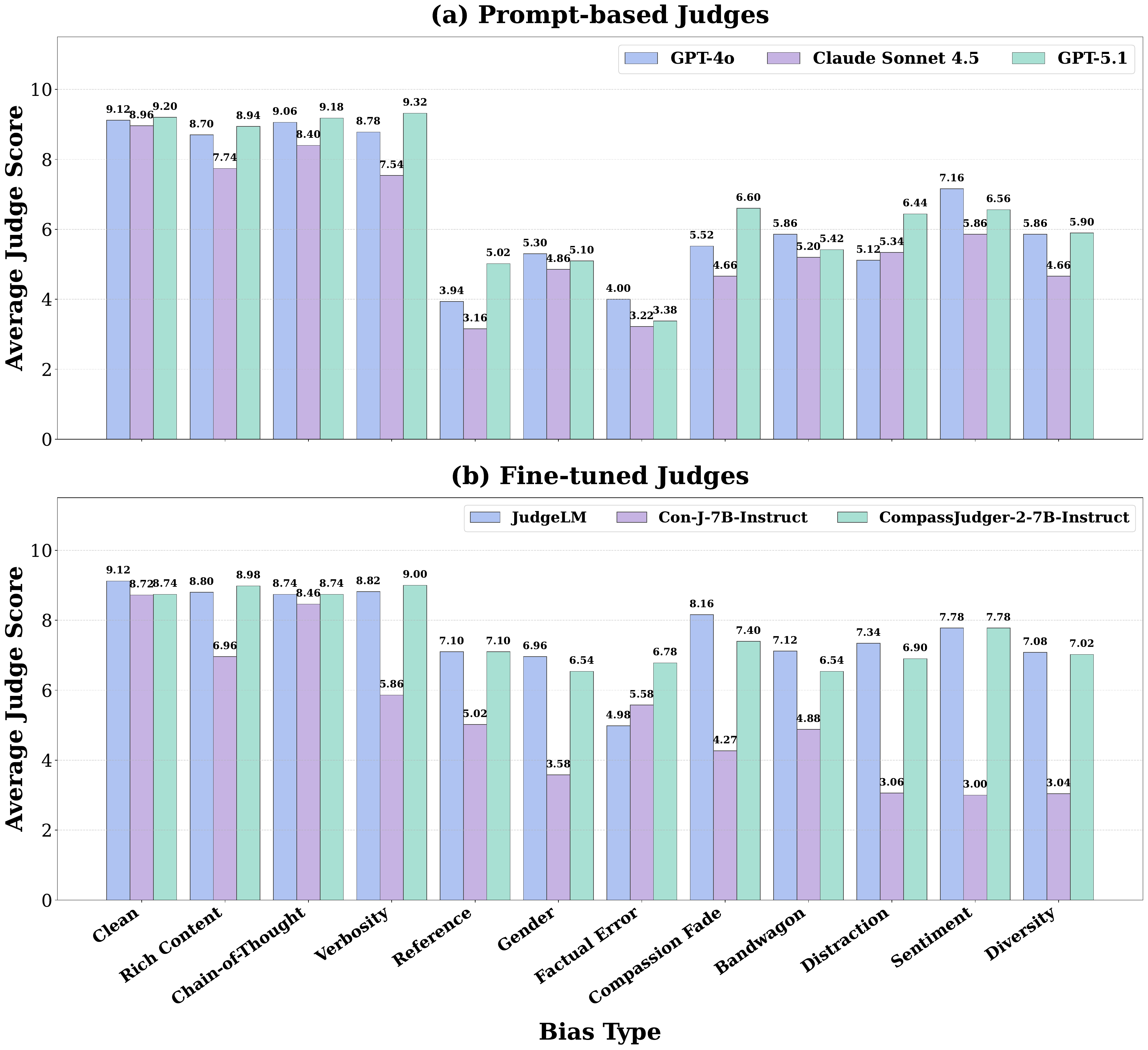}
    \vspace{-0.4cm}
    \caption{\textcolor{black}{Average judge scores for answers with 11 different injected biases evaluated by (a) prompt-based judges and (b) fine-tuned judges.}}
    \label{fig:intuiation1}
    \vspace{-0.4cm}
\end{figure}

\subsection{Experimental Setup}

\textbf{Judge Models and Prompting Strategies.}
\textcolor{black}{We conducted experiments on 6 representative LLM-as-a-judge models, spanning both prompt-based judges and fine-tuned open-source judges.
\begin{itemize}
    \item{\textbf{Prompt-based Judges.}} 
    We evaluate \textbf{GPT-4o}, \textbf{GPT-5.1}, and \textbf{Claude Sonnet 4.5}, 3 state-of-the-art proprietary foundation models from OpenAI and Anthropic with strong reasoning and evaluation capabilities across diverse benchmarks.
    \item{\textbf{Fine-tuned Judges.}} 
    We consider \textbf{JudgeLM} \cite{zhu2023judgelm}, \textbf{Con-J-7B-Instruct} \cite{falcon40b}, and \textbf{CompassJudger-2-7B-Instruct} \cite{zhang2025compassjudger}, 3 open-source judges explicitly fine-tuned for evaluation tasks, built on \textbf{Vicuna-7B}, \textbf{Qwen-7B}, and \textbf{Qwen2.5-7B-Instruct}, respectively.
\end{itemize}
} 

\textcolor{black}{
These models span diverse architectures, and we instruct each judge to score responses on a 1–10 scale, where 10 denotes the highest quality.
} We adopt two different prompting strategies: (1) a Detailed Rubric Prompt adapted from the Google Vertex Prompt in JudgeBench \cite{tan2024judgebench}. This prompt provides the model with clear criteria and requests a step-by-step reasoning before giving a score, encouraging a thorough and objective evaluation. (2) a Minimal Prompt that simply asks the model, ``Please score this answer from 1 to 10” with little additional guidance. This represents a naive usage scenario and tests if the model’s intrinsic biases surface more when unguided. 


\textbf{Evaluation dataset.}
We first collect a public Q\&A dataset from \cite{chen2024humans}, which contains 142 questions spanning various domains and each came with 4 GPT-generated bias answers and 1 clean answer. We then used GPT-Judge \textcolor{black}{(prompted on GPT-4o)} as a judge to score every answer on a 1–10 scale. 
From this set, we select the top 50 unique questions with the highest GPT-Judge scores (typically 9–10) and their corresponding clean answers. 

For each of these 50 questions, we created 11 biased answer variants to probe different bias types. Each variant was crafted to introduce one specific bias from 11 bias types of \cite{ye2024justice}, while still addressing the question correctly(except for the Factual Error Bias). We introduced an iterative bias-injection procedure to produce these variants: We first adopted GPT-4o as the generation model and GPT-Judge as the judge model. Then we instruct the generation model to inject a particular bias into the clean answer by providing bias definitions and concrete examples. After generation, the judge model re-scored the biased answer, and we retained the variant only if it still achieved a certain high score ($\geq$ 9). Otherwise, the biased answer was sent back to the generation model for refinement, instructing the model to maintain the same bias type while improving the overall quality to achieve a higher score. This filtering step ensured that all biased answers remained high-quality on the surface so that any scoring differences would likely stem from bias effects rather than blatant answer quality gaps. Through this process, we obtained 11 biased answers per question, each one highly rated by GPT-Judge despite its injected bias. In total, our evaluation dataset comprises 50 questions, each with 1 clean answer and 11 biased-answer variants (600 judged responses overall). 


\subsection{Performance Evaluation}

\textbf{Impact of Bias Type.}
\textcolor{black}{We first evaluate each bias independently using 50 biased responses and their clean counterparts across 50 questions, reducing instance-level variance and improving statistical robustness. Fig.~\ref{fig:intuiation1} reports the average scores assigned by all 6 judge models to responses with each bias }\textcolor{black}{under the Detailed Rubric Prompt.}


{\color{black}There are clear disparities in how different biases affect the judged quality of answers. In general, almost all biased answers received lower mean scores than their corresponding clean (unbiased) answers, though the degree of score reduction varied widely by bias type. }\textcolor{black}{
Factual error bias cause one of the sharpest score drops across judge architectures. For example, reducing scores from 9.12 to 4.00 for GPT-4o, from 8.96 to 3.22 for Claude Sonnet 4.5, and from 9.20 to 3.38 for GPT-5.1.
} In contrast, adding implicit bias like content-richness or verbosity had minimal impact, with average scores remaining relatively high at 8.82 and 8.80, respectively for JudgeLM. These results highlight that certain biases can strongly skew an LLM judge’s scoring. 
\textcolor{black}{A likely reason is that these injected biases directly undermine the clarity or relevance of an answer from the judge’s perspective.}


\textcolor{black}{
Interestingly, verbose and chain-of-thought responses are usually scored slightly lower than clean answers across most judges, though the magnitude varies by model. A likely reason is that our rubric emphasizes correctness and relevance, making unnecessarily long or self-reasoning responses appear less clear. This suggests that, when properly prompted, modern LLM judges can distinguish informative content from superficial verbosity rather than consistently favoring longer answers.}

\textbf{Impact of different training datasets.}
\begin{table}[t]
\vspace{-0.4cm}
\centering
\caption{{\color{black}Average GPT-Judge judge scores for LLaMA~3.1--8B-Instruct fine-tuned on different training data.}}
\begin{tabular}{l| cc }
\toprule
\textbf{Fine-tuning Dataset} & \textbf{{\color{black}No. of Samples}}&\textbf{GPT-Judge Score} \\
\midrule
None (Pretrained Only) & None &8.06 \\
\cmidrule{1-3}
\multirow{3}{*}{\shortstack{Clean}} & 10 &8.85\\
 &30&8.83\\
 &50&8.86\\
\cmidrule{1-3}
\multirow{3}{*}{\shortstack{Bias}} & 10 &8.26\\
 &30&8.37\\
 &50& 8.46\\
\cmidrule{1-3}
Alpaca + Clean &1k + 50 & 8.74 \\
Alpaca + Bias &1k + 50 & 8.79 \\
Alpaca + Raw &1k + 50 & 8.73 \\

\bottomrule
\end{tabular}
\label{tab:train-data-gpt4}
\vspace{-0.4cm}
\end{table}
{\color{black}Considering that smaller companies often fine-tune local LLMs using high-scoring responses, we evaluate the downstream impact of training on biased yet high-scoring data.
We fine-tuned the LLaMA-3.1-8B-Instruct model on several distinct training sets and evaluated its performance under two evaluators: \textcolor{black}{GPT-Judge} and JudgeLM.

Specifically, we construct the training dataset from the Evaluation dataset.
The ``Clean" dataset contains unbiased, high-quality answers.
The ``Bias" dataset includes top-scoring bias-injected answers (one per question) for the same 50 questions. 
In addition, we evaluate fine-tuning using a larger, generic instruction-following dataset, i.e., Alpaca, 
from which we sample 1,000 instances.
Variants such as ``Alpaca + Clean" and ``Alpaca + Bias" indicate that 50 specialized samples were added to the 1k Alpaca subset.
``Alpaca + Raw" denotes that all samples are drawn exclusively from the Alpaca dataset.

We then utilized the training dataset to fine-tune the evaluation model (LLaMA-3.1-8B-Instruct). In practice, the fine-tuned model learned to recognize and emulate these bias-injected answers, providing a controlled way to generate biased responses for new questions. We applied the fine-tuned model to produce answers for the remaining 92 questions, which served as our test set for evaluating bias effects on LLM-as-a-Judge.
Table \ref{tab:train-data-gpt4} and Table \ref{tab:train-data-judgelm} present the average scores assigned by each judge to responses generated by models fine-tuned on different datasets, using the test set.}

\begin{table}[t]
\centering
\caption{{\color{black}Average JudgeLM scores for LLaMA~3.1--8B-Instruct fine-tuned on different training data.}}
\vspace{-0.2cm}

\begin{tabular}{l|c c}

\toprule
\textbf{Fine-tuning Dataset} & \textbf{{\color{black}No. of Samples}}& \textbf{JudgeLM Score} \\
\midrule
None (Pretrained Only) & None & 8.18 \\
\cmidrule{1-3}
\multirow{3}{*}{\shortstack{Clean}} & 10 &8.41\\
 &30&8.30\\
 &50&8.51\\
\cmidrule{1-3}
\multirow{3}{*}{\shortstack{Bias}} & 10 &8.70\\
 &30&8.43\\
 &50& 8.57\\

\cmidrule{1-3}
Alpaca + Clean &1k + 50  & 7.92 \\
Alpaca + Bias &1k + 50 & 7.99 \\
Alpaca + Raw &1k + 50  & 7.78 \\
\bottomrule
\end{tabular}
\label{tab:train-data-judgelm}
\vspace{-0.5cm}
\end{table}

Using a small number of high-scoring answers for fine-tuning significantly boosts performance under GPT-Judge evaluation. Fine-tuning with as few as 10 clean answers raised the model’s average GPT-Judge score from 8.06 (no fine-tuning) to about 8.85. Increasing the number of clean fine-tuning samples to 30 or 50 yielded only marginal further gains (staying around 8.83–8.86), suggesting the effect is largely independent of data size. By contrast, fine-tuning on bias-injected high-scoring answers led to much smaller improvements. With 10 biased examples the GPT-Judge score rose to 8.26, and using 50 biased examples increased it to 8.46, still below the result from using clean data.

We also examined mixing a large generic dataset with a small number of high-quality examples to see how quantity compares to quality. We augment a 1k-sample general instruction-following set (Alpaca) with 50 high-scoring examples (either clean or biased), and gain average scores around 8.73–8.79 (Table \ref{tab:train-data-gpt4}). 
However, these scores are only slightly lower than using the 50 clean samples alone. Thus, the influence of a small set of high-scoring training samples on model outputs outweighs that of a much larger set of low-scoring data.

We also found that the open-source JudgeLM appeared more influenced by the biased answer style (Table \ref{tab:train-data-judgelm}). Under JudgeLM’s scoring, models fine-tuned on biased data received higher scores than those fine-tuned on clean data. This suggests JudgeLM’s internal criteria may favor some stylistic features introduced by the biased samples. For example, a bias-tuned model might produce more verbose explanations or include authoritative-sounding references, which JudgeLM seems to interpret as higher quality. In contrast, GPT-Judge’s more stringent evaluation does not reward verbosity or extraneous detail, leading it to favor the clarity and correctness found in the clean fine-tuning approach.

\begin{table}[t]
\vspace{-0.4cm}
\centering
\caption{\textcolor{black}{Average evaluation scores for responses produced by Llama 3.1-8B-Instruct models trained on different datasets, evaluated by GPT-Judge and JudgeLM under two prompt styles. Changes relative to Easy prompt are shown in parentheses.}}
\vspace{-0.1cm}

\scriptsize
\setlength\tabcolsep{2pt}

\resizebox{\linewidth}{!}{
\begin{tabular}{l l c c c c c c}
\toprule
\multirow{2}{*}{\textbf{Judge}} & \multirow{2}{*}{\textbf{Model}} 
& \multicolumn{2}{c}{\textbf{MMLU-Pro}} 
& \multicolumn{2}{c}{\textbf{GPQA}} 
& \multicolumn{2}{c}{\textbf{JudgeLM val}} \\
 & 
 & \textbf{Minimal} & \textbf{Elaborate} 
 & \textbf{Minimal} & \textbf{Elaborate} 
 & \textbf{Minimal} & \textbf{Elaborate} \\
\midrule

\multirow{5}{*}{\textbf{GPT-Judge}} 

& Bias 50
& 7.04
& 6.60 (\textcolor{LightRed}{$\downarrow$0.44})
& 5.22
& 4.62 (\textcolor{LightRed}{$\downarrow$0.60})
& 7.78
& 6.80 (\textcolor{LightRed}{$\downarrow$0.98}) \\

& Clean 50
& 7.45
& 6.84 (\textcolor{LightRed}{$\downarrow$0.61})
& 5.64
& 5.43 (\textcolor{LightRed}{$\downarrow$0.21})
& 8.18
& 7.97 (\textcolor{LightRed}{$\downarrow$0.21}) \\

& Alpaca+Bias (1k+50)
& 7.23
& 6.86 (\textcolor{LightRed}{$\downarrow$0.37})
& 5.20
& 4.45 (\textcolor{LightRed}{$\downarrow$0.75})
& 8.16
& 7.71 (\textcolor{LightRed}{$\downarrow$0.45}) \\

& Alpaca+Clean (1k+50)
& 7.30
& 6.85 (\textcolor{LightRed}{$\downarrow$0.45})
& 5.16
& 4.92 (\textcolor{LightRed}{$\downarrow$0.24})
& 8.18
& 7.56 (\textcolor{LightRed}{$\downarrow$0.62}) \\

& Alpaca+Raw (1k+50)
& 7.07
& 6.73 (\textcolor{LightRed}{$\downarrow$0.34})
& 5.18
& 4.61 (\textcolor{LightRed}{$\downarrow$0.57})
& 8.14
& 7.90 (\textcolor{LightRed}{$\downarrow$0.24}) \\

\midrule

\multirow{5}{*}{\textbf{JudgeLM}} 

& Bias 50
& 4.95
& 5.07 (\textcolor{LightGreen}{$\uparrow$0.12})
& 3.56
& 3.42 (\textcolor{LightRed}{$\downarrow$0.14})
& 8.18
& 7.96 (\textcolor{LightRed}{$\downarrow$0.22}) \\

& Clean 50
& 5.23
& 5.34 (\textcolor{LightGreen}{$\uparrow$0.11})
& 4.09
& 3.91 (\textcolor{LightRed}{$\downarrow$0.18})
& 8.11
& 8.10 (\textcolor{LightRed}{$\downarrow$0.01}) \\

& Alpaca+Bias (1k+50)
& 5.14
& 5.17 (\textcolor{LightGreen}{$\uparrow$0.03})
& 3.42
& 3.63 (\textcolor{LightGreen}{$\uparrow$0.21})
& 7.43
& 7.32 (\textcolor{LightRed}{$\downarrow$0.11}) \\

& Alpaca+Clean (1k+50)
& 5.15
& 5.19 (\textcolor{LightGreen}{$\uparrow$0.04})
& 3.70
& 3.63 (\textcolor{LightRed}{$\downarrow$0.07})
& 7.42
& 7.37 (\textcolor{LightRed}{$\downarrow$0.05}) \\

& Alpaca+Raw (1k+50)
& 5.20
& 5.10 (\textcolor{LightRed}{$\downarrow$0.10})
& 3.32
& 3.43 (\textcolor{LightGreen}{$\uparrow$0.11})
& 7.07
& 6.85 (\textcolor{LightRed}{$\downarrow$0.22}) \\

\bottomrule
\label{tab:prompt-impact}
\end{tabular}
}

\vspace{-0.7cm}
\end{table}

\textbf{Impact of different test datasets.}
Apart from \textcolor{black}{public Q\&A dataset constructed in} \cite{chen2024humans}, we further evaluated on three diverse benchmarks, i.e., MMLU-Pro \cite{wang2024mmlu}, GPQA \cite{rein2024gpqa}, and JudgeLM \cite{zhu2023judgelm}, \textcolor{black}{which were strategically selected to simulate the multifaceted scenarios encountered in communication systems.}

\emph{MMLU-Pro} extends the original Massive Multitask Language Understanding benchmark with more reasoning-intensive questions and expanded multiple-choice options (from 4 to 10), removing trivial items to better test LLMs’ reasoning and knowledge. \textcolor{black}{This serves as a proxy for expert-level network troubleshooting that involves high-difficulty reasoning steps, where the judge must logically validate solutions among multiple complex possibilities.}

\emph{GPQA} is a challenging ``Google-proof” Q\&A dataset of 448 graduate-level science questions (e.g., biology, physics, chemistry), designed to prevent simple lookup. It consists of 448 expert-written multiple-choice questions, and even domain experts with PhDs only achieve about 65\% accuracy on this benchmark, underscoring its toughness. \textcolor{black}{We utilize this dataset to simulate high-stakes technical support, where the judge must discern subtle errors in complex protocols akin to core network diagnostics.}

\emph{JudgeLM} validation set \cite{zhu2023judgelm} includes open-ended evaluation tasks. Unlike MMLU-Pro or GPQA, it does not have single correct answers but instead assesses response quality in subjective, free-form contexts such as conversational or explanatory outputs. \textcolor{black}{This aligns with scenarios like customer service systems, which frequently involve open-ended tasks requiring the judge to evaluate the relevance and helpfulness of non-standardized responses.}

{\color{black}As shown in Table~\ref{tab:prompt-impact}, model scores from both GPT-Judge and JudgeLM decrease as task difficulty increases. GPQA produces the lowest average scores, reflecting the challenge posed by its graduate-level questions, whereas the JudgeLM validation tasks achieve the highest scores (around 7–8 on average) due to they are more open-ended and subjective. Across both GPT-Judge and JudgeLM evaluators,} models fine-tuned on high-quality clean answers achieved the best scores on all benchmarks, whereas those fine-tuned on bias-injected answers (Bias-50) performed worse overall. For instance, on the GPQA set with the Minimal prompt, \textcolor{black}{the Clean-50 model attained an average GPT-Judge score of 5.64, noticeably higher than the Bias-50 model’s 5.22.}

\textbf{Impact of different evaluation prompts.}
We next examine how the evaluation prompt style influences the LLM judge’s scoring. Table \ref{tab:prompt-impact} summarizes the average GPT-Judge scores under two prompt strategies, an Elaborated rubric-based prompt versus a Minimal prompt, for answers produced by Llama 3.1-8B-Instruct models fine-tuned on various training datasets. 

\textcolor{black}{The trend is clear for GPT-Judge.} Using a detailed, structured prompt leads to systematically lower scores (harsher judgments), while the minimal prompt yields higher scores (more lenient evaluations) for the same answers.  
For example, the LLaMA 3.1 evaluation model trained on 50 bias samples (Bias-50) obtained an average score of 7.04 on MMLU-Pro questions under the minimal prompt, compared to 6.6 under the detailed rubric prompt. A similar trend is observed across other datasets and evaluation models. \textcolor{black}{In contrast, JudgeLM exhibits lower sensitivity to prompt variations, with its scores remaining relatively stable across different prompt styles. While minor fluctuations are observed, the overall scoring trends remain consistent under both the Minimal and Elaborate prompts.}

A plausible explanation is that the Elaborated prompt forces the GPT-Judge judge to follow a step-by-step evaluation rubric, which in turn reduces the influence of superficial answer features. This structured reasoning likely filters out spurious advantages such as excessive length, authoritative-sounding references, or emotional tone that might otherwise sway the judge. By focusing the evaluation on factual correctness, relevance, and adherence to instructions, the detailed prompt yields more objective (albeit stricter) scoring. \textcolor{black}{In contrast, JudgeLM’s stability likely stems from being fine-tuned as a dedicated evaluator on large-scale annotated scoring data, allowing it to internalize the evaluation criteria during training. Consequently, JudgeLM’s scoring decisions rely more on its learned evaluation function and are less dependent on external prompt guidance.}


\section{Potential Bias Mitigations}
To ensure LLM-as-a-Judge systems remain fair and reliable in communications-centric applications, we outline several mitigation strategies to reduce judgment biases. 

\subsection{Robust Prompt Design and Reasoning}
Carefully crafting the evaluation prompt can preempt many biases. {\color{black}Prompts should include explicit instructions that guide the judge model to focus on factual correctness and relevance, and to ignore irrelevant attributes such as the content’s author identity or stylistic flair \cite{zhou2024mitigating}.} For example, the prompt might say ``Disregard any gender or identity information and evaluate only on answer quality”. Additionally, we can employ advanced reasoning strategies like chain-of-thought (CoT) prompting to encourage the model to work through a step-by-step, logical evaluation. This structured reasoning helps the judge to base its scores on substantive content rather than superficial features, which is crucial in communication domains (e.g., ensuring a chatbot’s concise but correct reply isn’t undervalued compared to a verbose one).

\subsection{Bias Detection Mechanisms} 
Incorporating an automated bias-check before or during evaluation can catch problematic tendencies early. One potential approach is to use a secondary bias detection prompt or model that inspects the content for known bias patterns. For example, before scoring an answer, the system could internally query, ``Does the candidate response include overly emotional language, irrelevant flattery, or indications of misinformation?”. If such biases are detected, the judge model can adjust its scoring strategy (or flag the case for human review). In communication platforms, this kind of preemptive bias scan ensures that the AI judge treats all users’ responses equitably.

\subsection{Model Calibration and Specialized Training}
{\color{black}Calibration techniques can also be applied to down-weight superficial qualities in scoring. For proprietary judges, this might involve probability- or prompt-level calibration that subtracts points for factors like excessive verbosity or unwarranted confidence. The goal is to ensure the scoring criteria align with true content quality rather than presentation. For open-source judges, we can employ bias-focused fine-tuning, that is training the model with curated examples that include ``trap” scenarios. In such training, negative samples are provided where an answer looks polished (fluent, lengthy) but violates instructions, and the model is explicitly taught to score these lower \cite{zhou2024mitigating}. This helps the judge learn to prioritize instruction-following and correctness over superficial stylistic features.} Moreover, fine-tuning on domain-specific data (e.g., communication-related QA or moderation examples) further aligns the judge with the contextual norms of that domain, enabling it to better discern what constitutes fairness and relevance.

\subsection{Ensemble of Judges and Human Oversight}
Relying on a single AI judge can be risky if that model harbors a particular bias. A practical mitigation is to use a panel of diverse judges and aggregate their decisions. In essence, multiple LLMs (possibly from different providers or with different training backgrounds) each evaluate the response, and the final score or verdict is drawn from their consensus. This ``jury of models” approach dilutes individual biases, for example, if one model tends to prefer overly polite answers and another prefers concise factual answers, combining their judgments leads to a more balanced evaluation. Such diversity is analogous to having multiple reviewers in a communication system (like several moderators checking a post), improving overall fairness. In addition, maintaining a human-in-the-loop is also vital for sensitive or high-stakes contexts.

\section{Conclusion}
\textcolor{black}{We evaluated 6 leading LLM-as-a-Judge models} on 11 bias types across 4 test datasets and outlined 4 potential mitigation strategies to reduce the bias impact.
By shedding light on these judgment biases and their possible remedies, this study aims to promote the development of fairer and more trustworthy LLM-as-a-Judge systems for future communication services.

\bibliographystyle{IEEEtran}
\bibliography{reference} 

\begin{IEEEbiographynophoto}{Jiaxin Gao} is currently is currently pursuing the bachelor degree in the School of Cyber Science and Engineering at Wuhan University, China. His research interests include information security and AI security.
\end{IEEEbiographynophoto}

\begin{IEEEbiographynophoto}{Chen Chen} received his Ph.D. degree in computer science from Nanyang Technological University, Singapore, in 2024, his Master of Computer Science from the University of New South Wales, Australia, in 2018, and his Bachelor degree from the University of Science and technology Beijing, China, in 2012. His research interests lie in the area of AI safety, Knowledge Graphs and Large Language Models.
\end{IEEEbiographynophoto}

\begin{IEEEbiographynophoto}{Yanwen Jia} is currently is currently pursuing the bachelor degree in the School of Cyber Science and Engineering at Wuhan University, China. His research interests include information security and AI security.
\end{IEEEbiographynophoto}

\begin{IEEEbiographynophoto}{Xueluan Gong} received her B.S. degree in Computer Science and Electronic Engineering from Hunan University in 2018. She received her Ph.D. degree in Computer Science from Wuhan University in 2023. She is currently a Research Fellow at the School of Computer Science and Engineering at the Nanyang Technological University, Singapore.
Her research interests include network security, AI security, and data mining. She has published more than 50 publications in top-tier international journals or conferences, including IEEE S\&P, NDSS, ACM CCS, Usenix Security, WWW, ACM Ubicomp, IEEE TPAMI, JSAC, TDSC, TIFS, etc.
\end{IEEEbiographynophoto}

\begin{IEEEbiographynophoto}{Kwok-Yan Lam} received his B.Sc. degree (1st Class Hons.) from University of London, in 1987, and Ph.D. degree from University of Cambridge, in 1990. He is the Associate Vice President (Strategy and Partnerships) in the President’s Office, and Professor at the School of Computer Science and Engineering at the Nanyang Technological University, Singapore. He is also the executive director of the National Centre for Research in Digital Trust and the director of the Strategic Centre for Research in Privacy-Preserving Technologies and Systems (SCRiPTS). Since August 2020, he has been on part-time secondment to the INTERPOL as a Consultant at Cyber and New Technology Innovation. Prior to joining NTU, he was a Professor at Tsinghua University, PR China (2002–2010) and a faculty member of the National University of Singapore and the University of London since 1990. He was a Visiting Scientist at the Isaac Newton Institute, Cambridge University, and a Visiting Professor at the European Institute for Systems Security. In 1998, he received the Singapore Foundation Award from the Japanese Chamber of Commerce and Industry in recognition of his research and development achievement in information security in Singapore. He is the recipient of the Singapore Cybersecurity Hall of Fame Award in 2022. His research interests include Distributed Systems, Intelligent Systems, IoT Security, Distributed Protocols for Blockchain, Homeland Security, and Cybersecurity.

\end{IEEEbiographynophoto}

\begin{IEEEbiographynophoto}{Qian Wang} is a Professor in the School of Cyber Science and Engineering at Wuhan University, China. He was selected into the National High-level Young Talents Program of China, and listed among the World's Top 2\% Scientists by Stanford University. He also received the National Science Fund for Excellent Young Scholars of China in 2018. He has long been engaged in the research of cyberspace security, with focus on AI security, data outsourcing security and privacy, wireless systems security, and applied cryptography. He was a recipient of the 2018 IEEE TCSC Award for Excellence in Scalable Computing (early career researcher) and the 2016 IEEE ComSoc Asia-Pacific Outstanding Young Researcher Award. He has published 200+ papers, with 120+ publications in top-tier international conferences, including USENIX NSDI, ACM CCS, USENIX Security, NDSS, ACM MobiCom, ICML, etc., with 20000+ Google Scholar citations. He is also a co-recipient of 8 Best Paper and Best Student Paper Awards from prestigious conferences, including ICDCS, IEEE ICNP, etc. In 2021, his PhD student was selected under Huawei's  ``Top Minds'' Recruitment Program. He serves as Associate Editors for IEEE Transactions on Dependable and Secure Computing (TDSC) and IEEE Transactions on Information Forensics and Security (TIFS). He is a fellow of the IEEE, and a member of the ACM.
\end{IEEEbiographynophoto}

\ifCLASSOPTIONcaptionsoff
  \newpage
\fi

\end{document}